\documentclass[11pt,a4paper]{article}
\usepackage[hyperref]{emnlp2020}
\usepackage{times}
\usepackage{latexsym}

\usepackage{microtype}
\usepackage{url}
\usepackage{microtype}
\usepackage{helvet}
\usepackage{courier}
\usepackage{stfloats}
\usepackage{color}
\usepackage{amsmath}
\usepackage{amssymb}
\usepackage{array}
\usepackage{graphicx}
\usepackage{subfigure}
\usepackage{colortbl} 
\usepackage{arydshln} 
\usepackage{multirow} 
\usepackage{multicol}
\usepackage{wrapfig,lipsum,booktabs}
\usepackage[encapsulated]{CJK}
\usepackage{float}
\usepackage{mathrsfs}
\usepackage{mathtools}
\usepackage{amsopn}
\usepackage{url}
\usepackage{bm}
\usepackage{soul}
\usepackage{pgfplots}
\pgfplotsset{compat=1.14}
\DeclareMathOperator*{\argmax}{arg\,max}

\aclfinalcopy 

\title{Explicit Reordering for Neural Machine Translation}

\author{Kehai Chen, Rui Wang, Masao Utiyama, and Eiichiro Sumita \\
	National Institute of Information and Communications Technology (NICT), Kyoto, Japan \\
	\texttt{\{khchen, wangrui, mutiyama, eiichiro.sumita\}@nict.go.jp} \\
\\}

\date{}

\begin{document}
\maketitle
\begin{abstract}
	In Transformer-based neural machine translation (NMT), the positional encoding mechanism helps the self-attention networks to learn the source representation with order dependency, which makes the Transformer-based NMT achieve state-of-the-art results for various translation tasks.
	However, Transformer-based NMT only adds representations of positions sequentially to word vectors in the input sentence and does not explicitly consider reordering information in this sentence.
	In this paper, we first empirically investigate the relationship between source reordering information and translation performance.
	The empirical findings show that the source input with the target order learned from the bilingual parallel dataset can substantially improve translation performance.
	Thus, we propose a novel reordering method to explicitly model this reordering information for the Transformer-based NMT.
	The empirical results on the WMT14 English-to-German, WAT ASPEC Japanese-to-English, and WMT17 Chinese-to-English translation tasks show the effectiveness of the proposed approach.
\end{abstract}

\section{Introduction}
\label{1_Intro}
The positional encoding mechanism plays a very important role on the Transformer-based neural machine translation (NMT) system~\cite{NIPS2017_7181}.
Typically, it solely relies on the positions of words to learn positional embeddings to encode the order of dependencies between words in a sentence instead of the traditional recurrent~\cite{bahdanau2015neural} and convolutional~\cite{sutskever2014sequence} neural networks.
This allows the Transformer-based NMT to perform (multihead) and stack (multi-layer) attentive functions in parallel to learn the source representation with order information, which achieves state-of-the-art results for various translation tasks~\cite{barrault-etal-2019-findings}.

However, Transformer-based NMT only adds these learned positional embeddings sequentially to word vectors in the input sentence and does not explicitly consider reordering information between words in this sentence.
To address this issue, \newcite{chen-etal-2019-neural} attempted to implicitly penalize the given positional embedding to capture left-reordering information through a penalty vector each value of which has between zero and one. 
In other words, their approach implicitly modeled reordering information instead of explicitly reordering word positions which is widely used in phrase-based statistical machine translation (SMT)~\cite{P06-1066,D08-1089}, 
Thus, we hypothesize that explicitly modeling global reordering information is more straightforward and can help Transformer-based NMT model learn source reordering information to improve translation performance.

In this paper, we first empirically investigate the relationship between source reordering information and translation performance.
Our empirical finding shows that the source input, whose word order is in line with that of words in target language sentence, can substantially improve translation performance.
Based on this finding, we hypothesize that this target language order information in the source sentence can enhance the performance of the Transformer-based NMT model.
Thus, we propose a novel method to explicitly capture reordering information under the supervision of the source input with the target language order.
The empirical results on the WMT14 English-to-German, WAT ASPEC Japanese-to-English, and WMT17 Chinese-to-English tasks show that the proposed method gains significantly improvement over the strong baseline Transformer-based NMT model.

\section{Background:Transformer-based NMT}
\label{sec2_Background}
\subsection{Positional Encoding}
\label{sec2_1}
In Transformer-based NMT~\cite{NIPS2017_7181}, the positional encoding is used to capture ordering dependency between words in a sentence.
Formally, given a sequence of word vectors \textbf{\textit{X}}=\{$\textbf{x}_{1}, \cdots, \textbf{x}_{J}$\} for the input sentence, the positional embedding of each word is computed initially based on its position:
\begin{equation}
	\begin{split}
		& \textbf{pe}_{(pos,2k)}=\textup{sin}(pos/10000^{2k/d_{model}}),\\
		& \textbf{pe}_{(pos,2k+1)}=\textup{cos}(pos/10000^{2k/d_{model}}),
		\label{eq1:position_embedding}
	\end{split}
\end{equation}
where \(pos\) is the word's position index in the sentence, \(k\) is the dimension of the position index, and \(d_{model}\) is the dimension of word vector.
As a result, there is a sequence of positional embeddings \(\textbf{PE}=\{\textbf{pe}_{1}, \cdots, \textbf{pe}_{J}\)\}.
The word vector $\textbf{x}_{j}$ is then added with its $\textbf{pe}_{j}$ to yield a combined embedding $\textbf{v}_{j}$.
\(\textbf{H}^{0}\)=\{$\textbf{v}_{1}, \cdots, \textbf{v}_{J}\}$ serves as the input of Transformer-based NMT model to learn the source representation.

\subsection{Transformer}
\label{sec2_2}
For the Transformer-based NMT, the encoder is composed of a stack of $N$ identical layers, each of which includes two sub-layers.
The first sub-layer is a self-attention module, and the second sub-layer is a position-wise fully connected feed-forward network.
A residual connection~\cite{He2016DeepRL} is applied between the two sub-layers, and then layer normalization~\cite{DBLP:journals/corr/BaKH16} is performed.
Formally, the encoder is organized as follows to learn the source representation:
\begin{equation}
\begin{split}
&\textbf{C}^{n} = \textup{LN}(\textup{SelfATT}(\textbf{H}^{n-1})+\textbf{H}^{n-1}),\\
& \textbf{H}^{n} =\textup{LN}(\textup{FFN}(\textbf{C}^{n})+\textbf{C}^{n}),
\label{eq2:Encoder}
\end{split}
\end{equation}
where $\textup{SelfATT}(\cdot)$, $\textup{LN}(\cdot)$, and $\textup{FFN}(\cdot)$ are self-attention module, layer normalization, and feed-forward network for the $n$-th layer, respectively.

Similarly, the decoder is also composed of a stack of $N$ identical layers, in which there is an additional attention sub-layer between $\textup{SelfATT}(\cdot)$ and $\textup{FFN}(\cdot)$ in Eq.\eqref{eq2:Encoder} to compute alignment weights for the output of the encoder stack $\textbf{H}^{N}$.
Finally, the output of the stacked decoder is used to generate the target translation word-by-word.
To obtain an available translation model, the training objection maximizes the conditional probabilities over a parallel training corpus $\{[\textbf{X}, \textbf{Y}]\}$:
\begin{equation}
	\mathcal{J}(\theta)=\argmax_{\theta}\{P(\textbf{Y}|\textbf{X}; \theta)\}.
	\label{eq3:NMT_Training}
\end{equation}

\section{Preliminary Experiments}
\label{sec3_PreliminaryExperiment}
Machine translation primarily consists of two interrelated problems: predicting the words in the translation and determining on their order.
In particular, the word order of the generated translation should be consistent with the grammar of the target language as much as possible.
Actually, there are many distant language pairs with large differences in word order, such as Chinese-to-English~\cite{galley-manning-2008-simple} and Japanese-to-English translations~\cite{goto-etal-2013-distortion}.
In traditional SMT, many reordering methods that include the reordering model~\cite{P06-1066,D08-1089}, pre-ordering model~\cite{kawara-etal-2018-recursive}, and post-ordering model~\cite{goto-etal-2012-post} have been proposed to learn large-scale reordering rules and features to narrow the order differences between language pairs.
However, these reordering methods in SMT are not easy to compatible with the sequence-to-sequence NMT model.

\begin{figure*}[t]
		\begin{minipage}[b]{0.48\linewidth}
			\begin{center}
				\includegraphics[width=3.0in, height=1.35in]{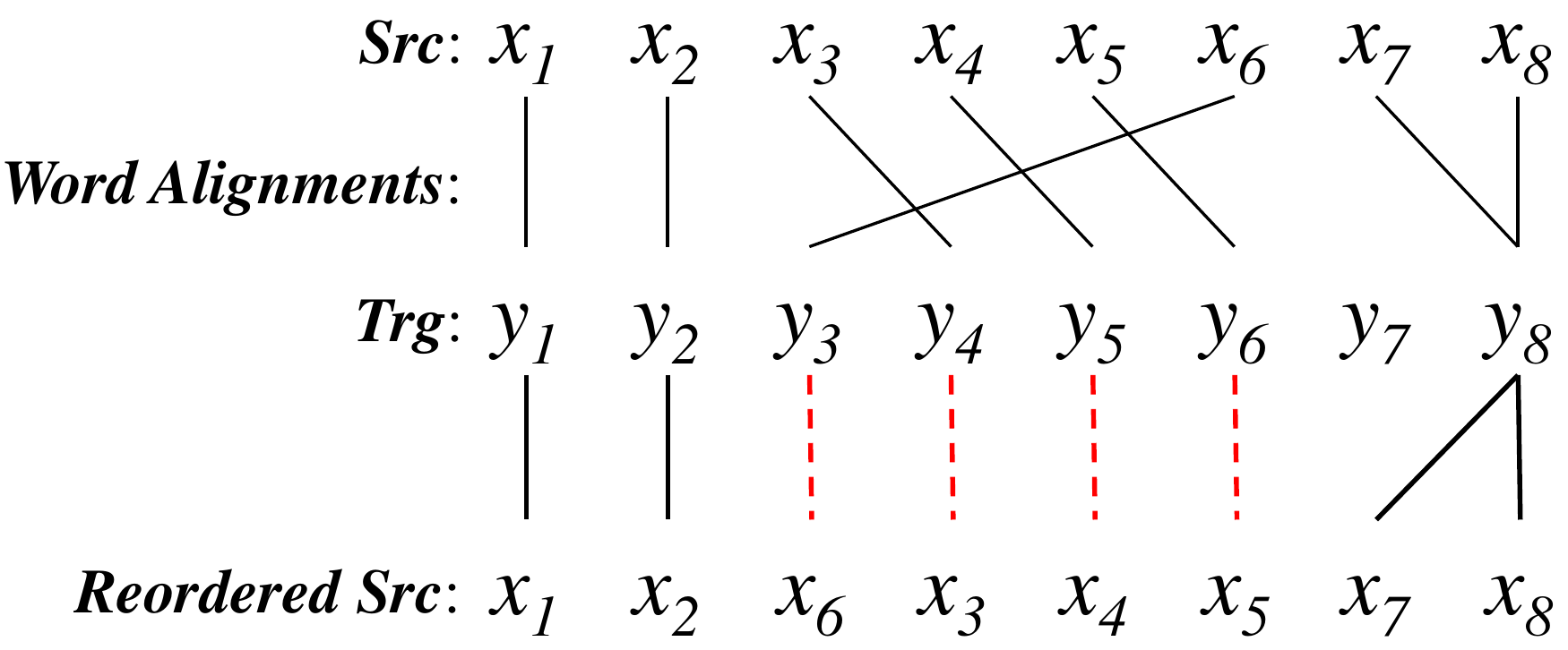}
			\end{center}
				\caption{A bilingual parallel sequence with word alignments and the corresponding reordered source sequence. The red dotted lines indicate that the positions of these source words are adjusted according to the target word order. }
		\label{fig1:WordAlignment}
		\end{minipage}
\hspace{0.1cm}
		\begin{minipage}[b]{0.48\linewidth}
			\begin{center}
				\pgfplotsset{height=5cm,width=7.0cm,compat=1.14,every axis/.append style={thick},legend columns=2}
				\begin{tikzpicture}
				\centering
				\begin{axis}[
				axis y line*=left,
				axis x line*=left,
				ybar=5pt,
				bar width=8pt,
				enlargelimits=0.15,
				legend style={at={(0.5,1.35)},
					anchor=north,legend columns=1},
				ylabel={BLEU},
				font=\small,
				symbolic x coords={En-De, Ja-En, Zh-En},
				xtick=data,
				bar width=9pt,
				nodes near coords,
				nodes near coords align={vertical},
				],
				\addplot+ coordinates {(En-De, 27.48) (Ja-En,31.1) (Zh-En,24.28)};
				\addplot+ coordinates {(En-De, 31.5) (Ja-En,34.9) (Zh-En,27.37)};
				\legend{\small Original training and test data\;\;\;, Reordered training and test data}
				\end{axis}
				\end{tikzpicture}
			\end{center}
			\caption{Results of the Transformer (base) model for the three translation tasks. \textbf{Note that the reordered test data was also learned from parallel test data.}}
		    \label{fig2:PreliminaryExperiments}
		\end{minipage}
\end{figure*}

To explore the reordering mechanism for NMT, we empirically chose three language pairs (distant language pairs: Chinese-English and Japanese-English; similar language pair: English-German) as the corpora.
Specifically, we first gained word alignments for bilingual parallel sentence pairs (including the training dataset, validation dataset, and test dataset) using a classical word alignment toolkit \textit{fast\_align}\footnote{https://github.com/clab/fast\_align} \cite{dyer-etal-2013-simple}.
We then reordered all source sentences into those with the target language order depending on the word alignments, as shown in Figure~\ref{fig1:WordAlignment}.
Based on the reordered training dataset, we used the state-of-the-art Transformer-based framework~\cite{NIPS2017_7181} to train an NMT model, and evaluated it over the reordered test dataset.
Figure~\ref{fig2:PreliminaryExperiments} shows the translation results of Transformer (base) models for the WMT14 English-to-German (En-De), WAT ASPEC Japanese-to-English (Ja-En), and the WMT17 Chinese-to-English (Zh-En) translation tasks. 
The detailed setting are shown in Section \ref{sec:data}. 
In Figure~\ref{fig2:PreliminaryExperiments}, when both training dataset and test dataset are reordered (please note that the reordered test data is learned from parallel test data with word alignments), NMT achieved a performance improvement of up to 4 BLEU points over three language pairs.
Thus, these empirical findings show the following:
\begin{itemize}
	\item The source input with the target order learned from the bilingual parallel dataset can substantially improve translation performance.
	\item The Transformer-based NMT is very sensitive to the order information of the source input.
\end{itemize}

\section{Explicit Reordering Methods}
Based on previous empirical findings, we propose two novel reordering methods, i.e, explicit global reordering and reordering fusion based source representation, that enable the Transformer-based NMT model to explicitly model this useful target language order in the source sentence.

Formally, given an aligned bilingual sentence pair \(\{X=x_{1}^{J}, Y=y_{1}^{I}, A\}\), where \(J\) and \(I\) denote the length of the source and target sentences and \(A=\{a_{1}, a_{2}, \cdots, a_{I}\}\) denotes the word alignment between the source sentence \(X=\{x_{1}, x_{2}, \cdots, x_{J}\}\) and target sentence \(Y=\{y_{1}, y_{2}, \cdots, y_{I}\}\).
We base on the word alignment \(A\) to learn a sequence of positions \(R=\{r_{1}, r_{2}, \cdots, r_{J}\}\) for the original \(X\) depending on the word order of \(Y\).\footnote{In fact, the word alignment \(A\) is learned over \(X=y_{1}^{I}, Y=x_{1}^{J}\) so that the converted \(R\) has the same length as the original \(X\). Meanwhile, we remain the positions of unaligned source words in \(R\).}
Finally, the reordered position \(r_j\) is considered as the input to the Eq.\eqref{eq1:position_embedding} to learn the reordered positional embedding \(\textbf{re}_j\):
\begin{equation}
	\begin{split}
		& \textbf{re}_{(r_j,2k)}=\textup{sin}(r_j/10000^{2k/d_{model}}),\\
		& \textbf{re}_{(r_j,2k+1)}=\textup{cos}(r_j/10000^{2k/d_{model}}).
		\label{eq4:position_embedding}
	\end{split}
\end{equation}
As a result, there is a sequence of reordered positional embeddings \(\textbf{RE}=\{\textbf{re}_1,\textbf{re}_2, \cdots, \textbf{re}_J\}\).
\textbf{During the training of NMT, the \(\textbf{RE}\) will be used as a supervised signal to guild the learning of reordering information. During the decoding, our NMT model can jointly learn target order information in the source sentence and translation, that is, the alignment information is not necessary.}
\begin{figure}[h]
	\centering
\includegraphics[width=2.3in, height=1.8in]{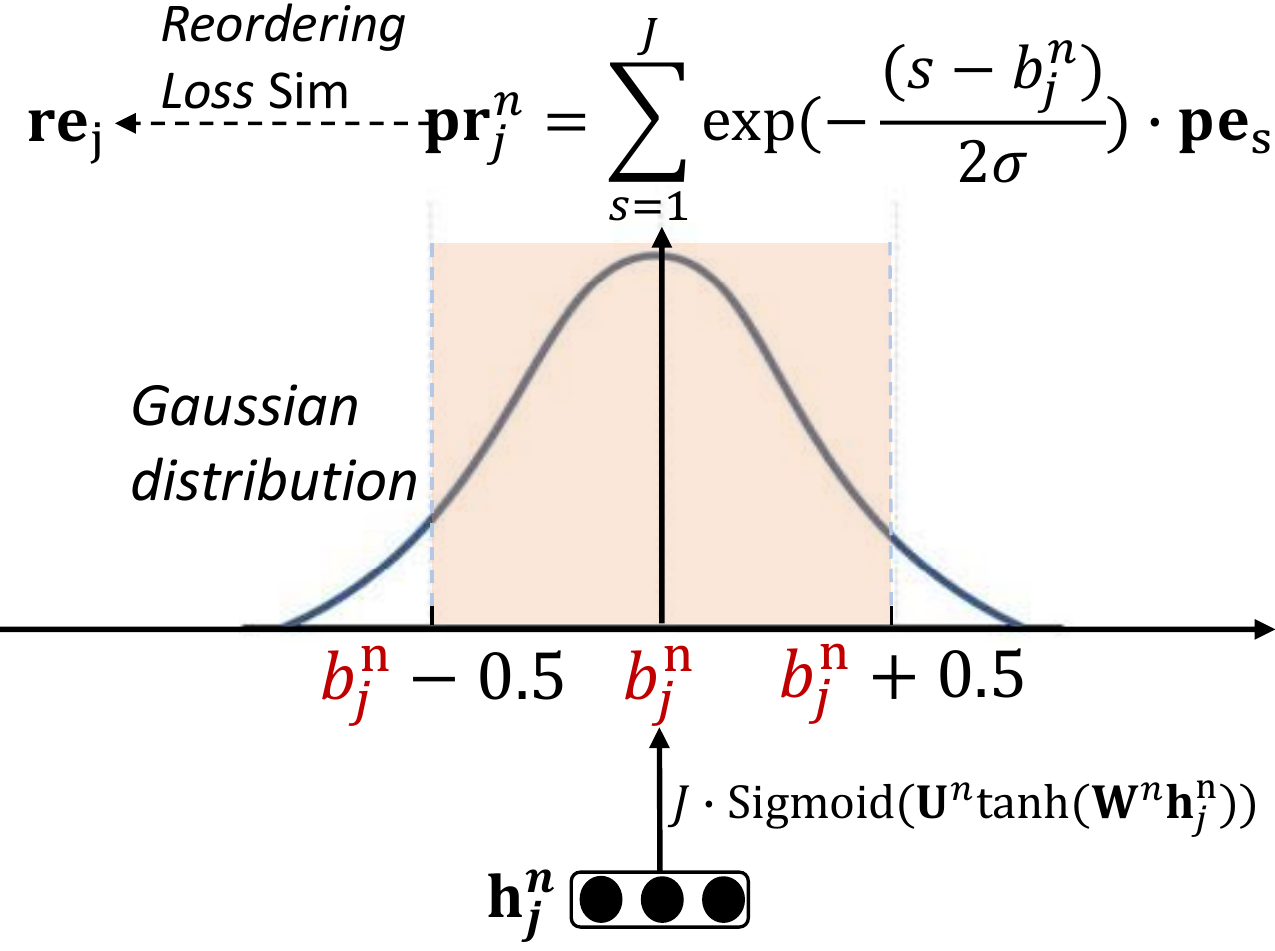}
\caption{The illustration of explicit global reordering.}
\label{fig3:ExGRE}
\end{figure}

\subsection{Method1: Explicit Global Reordering}
The proposed explicit global reordering (\textbf{ExGRE}) explicitly reorders the positions of words in the source sentence under the supervision of the target language order information. 
Moreover, this approach has another advantage of allowing the position of word to be reordered inside the entire sentence instead of to be only left-reordered~\cite{chen-etal-2019-neural}.
Specifically, a reordered position for one word \(x_j\) is first generated depending on its current context information.
We select a positional embedding of \(x_j\) from the existing positional embeddings by a Gaussian distribution centered on the reordered position \(b_{j}^{n}\) within the fixed window.

Formally, our model predicts its reordered position \(b_{j}^{n}\) for word \(x_j\) in the source sentence as follows:
\begin{equation}
	b_{j}^{n}=J \cdot \textup{sigmoid}(\textbf{U}^{n}\textup{tanh}(\textbf{W}^{n}\textbf{h}_{j}^{n})), \\
	\label{eq5:Position_Penalty_Scalar}
\end{equation}
where $\textbf{W}^{n}$$\in$$\mathbb{R}^{d_{model}\times 1}$ and $\textbf{U}^{n}$$\in$$\mathbb{R}^{d_{model}\times 1}$ are the parameters of model, and \(\textbf{h}_{j}^{n}\) is the \(j\)-th word's hidden state in the output \(\textbf{H}^{n}\) of the \(n\)-th layer in the encoder.
As a result of applying the sigmoid function, \(b_{j}^{n}\) \(\in [0, J]\).
Inspired by the work of \cite{luong-etal-2015-effective}, we place a Gaussian distribution centered around \(b_{j}^{n}\) to gain the reordered positional embeddings (See Figure~\ref{fig3:ExGRE}):
\begin{equation}
	\textbf{pr}_{j}^{n}=\sum_{s=1}^{J}\textbf{pe}_s \cdot \textup{exp}(-\frac{(s-b_{j}^{n})^{2}}{2\sigma}),
	\label{eq7:Reordered_Embeding}
\end{equation}
where the standard deviation is empirically set as \(\frac{D}{2}\), and \(D\) is empirically set to 0.5.
In particular, \(s\) is an integer within the windows [\(b_{j}^{n}\)-D, \(b_{j}^{n}\)+D] to ensure that the Gaussian distribution can give a max probability of our expected \(\textbf{pe}_s\) in the existing \textbf{PE}.
In contrast, Gaussian distribution gives minimal or zero probabilities of other positions outside the window [\(b_{j}^{n}\)-D, \(b_{j}^{n}\)+D].
The obtained \(\textbf{pr}_{j}^{n}\) is then added to the current \(\textbf{h}_{j}^{n}\) to reorder the positional of word \(x_j\):
\begin{equation}
	\overline{\textbf{h}}_{j}^{n}=\textbf{h}_{j}^{n} + \textbf{pr}_{j}^{n}.
	\label{eq7:Perform_Reordering}
\end{equation}
Naturally, the stacked encoder in Eq.\eqref{eq2:Encoder} is modified as follows to learn a reordering aware source representation \(\overline{\bm{\mathcal{H}}}^{N}\):
where \(B^{n}\) is \{\(b_{1}^{n}, \cdots, b_{J}^{n}\)\} and \(\textbf{PR}^{n}\) is \{\(\textbf{pr}_{1}^{n}, \cdots, \textbf{pr}_{J}^{n}\)\}.
\begin{equation}
\begin{split}
& \bm{\mathcal{C}}^{n} = \textup{LN}(\textup{SelfATT}(\overline{\bm{\mathcal{H}}}^{n-1})+\overline{\bm{\mathcal{H}}}^{n-1}),\\
& \bm{\mathcal{H}}^{n} =\textup{LN}(\textup{FFN}(\bm{\mathcal{C}}^{n})+\bm{\mathcal{C}}^{n}), \\
& B^{n}=J \cdot \textup{sigmoid}(\textbf{v}^{n}\textup{tanh}(\textbf{W}^{n}\bm{\mathcal{H}}^{n})), \\
& \textbf{PR}^{n}=\textbf{PE} \cdot \textup{exp}(-\frac{(s-B^{n})^{2}}{2\sigma}), \\
& \overline{\bm{\mathcal{H}}}^{n}=\bm{\mathcal{H}}^{n} + \textbf{PR}^{n}, \\
\label{eq8:Encoder_Reordering}
\end{split}
\end{equation}

Furthermore, the target order information found in Section~\ref{sec3_PreliminaryExperiment} is used as a supervised signal to guide the training of NMT. 
Formally, an additional reordering loss term is introduced to measure the reordered position, which encourages the translation model to learn expected word orders:
\begin{equation}
	\mathcal{J}(\theta)=\argmax_{\theta}\{P(Y|X; \theta)+\lambda*\sum_{j=1}^{J}\textup{Sim}(\textbf{pr}_{j}^{N}, \textbf{re}_{j}; \theta)\},
	\label{eq9:NMT_NewTraining}   
\end{equation}
where \(\textup{Sim}(\cdot)\) denotes the cosine distance of the position representations between each reordered positional embedding \(\textbf{pr}^{N}_j\)\(\in\)\(\textbf{PR}^{N}\) and the supervised one \(\textbf{re}_j\).
\(\lambda\) is used to control the weight of the reordering loss and is empirically set to 0.6 in the experiment.
Compared with the work of \newcite{chen-etal-2019-neural}, the differences are as follows:
\begin{enumerate}
	\item Our method performs the reordering based on the length of source sentence, and thereby allows the encoder to model the global reordering instead of only the left-reordering.
	\item The learned \(B^{N}\) is the explicit reordered positions of all words in a source sentence instead of the implicit reordered embeddings.
	\item We empirically introduce an additional reordering loss term to supervise the process of learning reordering.
\end{enumerate}
\begin{figure}[h]
	\centering
    \includegraphics[width=2.6in, height=3.2in]{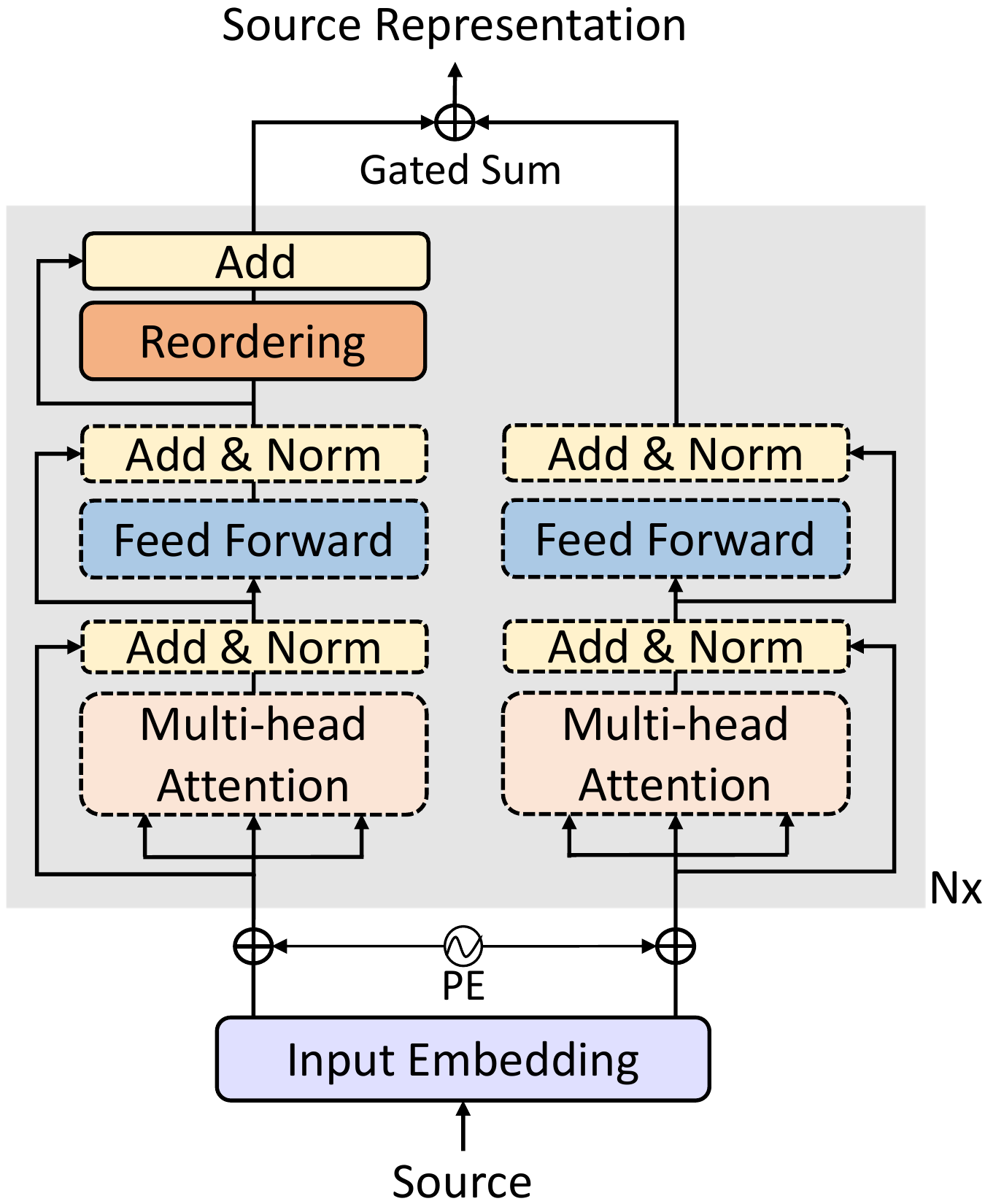}
    \caption{Proposed reordering fusion based source representation. Note that the dash modules are shared.}
    \label{fig4:ExFSR}
\end{figure}

\subsection{Method2: Reordering Fusion-based Source Representation}
According to the finding in Section~\ref{sec3_PreliminaryExperiment}, we believe that the source sentence with the original word order is more easily understood by humans whereas the source sentence with the target language order is more easily by the Transformer-based NMT model.
In other words, two types of word order are beneficial for the Transformer-based NMT model.
Thus, we propose a reordering fusion-based source representation (\textbf{ReFSR}) method to encode two types of word order information simultaneously into the final source representation to improve the performance of Transformer-based NMT model. 

Specifically, two types of source representation \(\textbf{H}^{N}\) and \(\overline{\bm{\mathcal{H}}}^{N}\) are learned based on Eqs.\eqref{eq2:Encoder} and \eqref{eq8:Encoder_Reordering}, respectively.
It is important to note that the SelfAtt and FNN modules are shared during the learning of \(\textbf{H}^{N}\) and \(\overline{\bm{\mathcal{H}}}^{N}\).
We then compute gate scalar $\lambda \in [0, 1]$ to weight the expected importance of two hidden states for different order information:
\begin{equation} g=\textup{sigmoid}(\textbf{U}_{g}\textbf{H}^{n}+\textbf{W}_{g}\overline{\bm{\mathcal{H}}}^{n}),
	\label{eq10:Gate_Scalar}
\end{equation}
where $\textbf{W}_{g}$$\in$$\mathbb{R}^{d_{model}\times 1}$ and $\textbf{U}_{g}$$\in$$\mathbb{R}^{d_{model}\times 1}$ are model parameters.
We then fuse $\textbf{H}^{N}$ and $\overline{\bm{\mathcal{H}}}$ to learn the final source representation:
\begin{equation} 
	\bm{\mathbb{H}}^{n}=g \cdot \overline{\bm{\mathcal{H}}}^{N}+(1-g) \cdot \textbf{H}^{N}.
	\label{eq11:Fusing_Sourece_Representation}
\end{equation}
Finally, \(\bm{\mathbb{H}}^{n}\) is fed to the decoder to learn a dependent-time context vector to predict the target translation word-by-word.
Note that there is a single aggregation layer to fuse two source representations with different order information.

\section{Experiments}
\subsection{Data}
\label{sec:data}
The proposed method was evaluated on three widely-used translation datasets: WMT14 En-De, WAT ASPEC Ja-En, and WMT17 Zh-En translation tasks which are standard corpora for NMT evaluation.

1) For the WMT14 En-De translation task, 4.43M bilingual sentence pairs from the WMT14 dataset were used as the training data.
The \textit{newstest2013} and \textit{newstest2014} datasets were used as the validation set and test set, respectively.

2) For the WAT ASPEC Ja-En translation task, 2M bilingual sentence pairs from the ASPEC corpus~\cite{nakazawa-etal-2016-aspec} were used as the training dataset. 
The validation set consists of 1,790 sentence pairs and the test set of 1,812 sentence pairs.

3) For the WMT17 Zh-En translation task, 22M bilingual sentence pairs from the WMT17 dataset were used as training data.
The \textit{newsdev2017} and \textit{newstest2017} datasets were used as the validation set and the test set, respectively. 

Besides, we applied our method to fully aligned six language parallel United Nation corpus\footnote{https://conferences.unite.un.org/UNCORPUS/en} to evaluate the effect of different language pairs.
The training data of each language pair included 11.3M bilingual sentence pairs. The validation set consists of 4,000 sentence pairs and the test set of 4,000 sentence pairs.

\subsection{Baseline Systems}
In addition to a vanilla Transformer base/big (\textbf{Trans.base/big}) models~\cite{NIPS2017_7181}, other comparison systems were as follows:

\textbf{+Relative PE}: incorporates relative positional embeddings into the self-attention mechanism of the Transformer-based NMT model~\cite{shaw-etal-2018-self}.

\textbf{+Reordering Embedding}: proposes a reordering mechanism to penalize the given positional embedding of a word based on its contextual information to implicitly capture (left-)reordering information~\cite{chen-etal-2019-neural}.

\textbf{+Recurrent PE}: proposes a recurrent positional embedding approach based on the
word vector instead of a static position index, and thereby encodes word content-based order dependencies into the source representation~\cite{chen-etal-2019-recurrent}.

\textbf{+Structural PE}: uses an additional dependency tree to represent the grammatical structure of a sentence, and encodes positional relationships among words as structural position representations to model the latent structure of
the input sentence~\cite{wang-etal-2019-self}.

Additionally, we report some recent impressive results for the WMT17 Zh-En translation task, for example, convolutional self-attention networks (\textbf{CSANs}) \cite{yang-etal-2019-convolutional} and the bi-attentive recurrent network (\textbf{BIARN}) \cite{hao-etal-2019-modeling} for Transformer-based NMT.
\subsection{System Setting}
The byte pair encoding algorithm~\cite{sennrich-haddow-birch:2016:P16-12} was adopted, and the vocabulary size was set to 40K. 
The dimension of all input and output layers was set to 512, and that of the inner feedforward neural network layer was set to 2048. 
The total heads of all multi-head modules was set to 8 in both the encoder and decoder layers. 
In each training batch, there was a set of sentence pairs containing approximately 4096\(\times\)8 source tokens and 4096\(\times\)8 target tokens. 
During training, the value of label smoothing was set to 0.1, and the attention dropout and residual dropout were \textit{p} = 0.1. 
The learning rate was varied under a warm-up strategy with warmup steps of 8,000.
For evaluating the test sets, following the training of 300,000 batches, we used a single model obtained by averaging the last 5 checkpoints, which were validated the model with an interval of 2,000 batches on the dev set.
During the decoding, the beam size was set to 5. 
All models were trained and evaluated on eight V100 GPUs.
The multi-bleu.perl script was used as the evaluation metric for the three translation tasks, and signtest~\cite{collins-koehn-kucerova:2005:ACL} was used as the statistical significance test.
We implemented the proposed NMT models on the \textit{fairseq} toolkit~\cite{ott-etal-2019-fairseq}.

\subsection{Main Results}
\begin{table*}[t]
	\centering
	\scalebox{.80}{
		\begin{tabular}{l||l|r|r||l|r|l|r}
		\hline \hline
			\multicolumn{1}{c||}{\multirow{2}{*}{Architecture}} & \multicolumn{3}{c||}{En-De} & \multicolumn{2}{c|}{Ja-En} & \multicolumn{2}{l}{Zh-En} \\ \cline{2-8} 
			\multicolumn{1}{c||}{}                              & BLEU   & \#Speed.   & \#Para.   & BLEU        & \#Para.        & BLEU        & \#Para.        \\ \hline
			\hline
			\multicolumn{8}{c}{\textit{Existing NMT systems}}                                                                                                 \\ \hline
			Trans.base \cite{NIPS2017_7181}  & 27.3 & N/A & 65.0M  &   N/A   &   N/A      &  N/A   &  N/A \\ 
			\;\;+Relative PE~\cite{shaw-etal-2018-self} &  26.8      &     N/A    &   N/A      &      N/A       &       N/A       &      N/A       &    N/A          \\ 
			\;\;+Reordering PE~\cite{chen-etal-2019-neural} &  28.22      &  8.6k       &   106.8M      &   31.41          &      84.4M        &  N/A           &     N/A         \\ 
			\;\;+Recurrent PE~\cite{chen-etal-2019-recurrent} &  28.35      &  9.9k       &   97.72M      &     N/A        &   N/A           &   N/A          &   N/A           \\ 
			\;\;+CSANs~\cite{yang-etal-2019-convolutional} & 28.18 & N/A & 88.0M  &  N/A  &   N/A      &  24.80   &  N/A   \\
			\;\;+BIARN~\cite{hao-etal-2019-modeling} & 28.21 & N/A & 97.4M  &  N/A   &   N/A      & 24.70 & 107.3M  \\ \cdashline{1-8}
			Trans.big \cite{NIPS2017_7181}  & 28.4  & N/A  & 213.0M  &  N/A   &  N/A       &  N/A  &  N/A    \\ 
			\;\;+Relative PE~\cite{shaw-etal-2018-self} &   29.2     &  N/A      &   N/A      &      N/A       &  N/A            &       N/A      &   N/A           \\ 
			\;\;+Reordering PE~\cite{chen-etal-2019-neural} &  29.11      &  3.4k       &    308.2M     &    31.93         &   273.7M           &   N/A          &      N/A        \\ 
			\;\;+Recurrent PE~\cite{chen-etal-2019-recurrent} &  29.11      &    N/A     &   289.1M     &    N/A         &      N/A        &      N/A       &    N/A          \\ 
			\;\;+Structural PE~\cite{wang-etal-2019-self} &   28.88     &    N/A     &    N/A     &    N/A         &  N/A            &     N/A        &    N/A          \\ 
			\;\;+CSANs~\cite{yang-etal-2019-convolutional} & 28.74 & N/A & 339.6M  & N/A    &   N/A      & 25.01 & N/A    \\
			\;\;+BIARN~\cite{hao-etal-2019-modeling} & 28.98 & N/A & 333.5M  &  N/A   &   N/A      &  25.10 & 373.3M    \\ \hline \hline
			\multicolumn{8}{c}{\textit{Our NMT systems}}                                                                                                      \\ \hline
			Trans.base  & 27.57 & 13.2k & 66.5M  &  31.21   &   67.5M      &  24.29   &  74.7M \\ 
			\;\;+ExGRE &  28.44\(\Uparrow\) &  13.1k & 66.5M  &  31.96\(\Uparrow\)  &  67.5M       &  24.91\(\Uparrow\)   &  74.7M            \\ 
			\;\;+ReFSR &  28.55\(\Uparrow\) &  11.9k & 66.5M & 32.36\(\Uparrow\)  &  67.5M       &  25.13\(\Uparrow\)   &  74.7M            \\ \cdashline{1-8}
			Trans.big  & 28.59 & 11.2k & 221.2M  &  32.18   &   223.1M      &  24.72   &  237.5M \\ 
			\;\;+ExGRE &  29.28\(\Uparrow\) &  11.1k & 221.2M  &  32.83\(\Uparrow\)  &   223.1M      &  25.19\(\Uparrow\)   & 237.5M             \\ 
			\;\;+ReFSR &  29.47\(\Uparrow\) &  9.9k & 221.2M &   33.27\(\Uparrow\)    &  223.1M   &  25.34\(\Uparrow\)   & 237.5M         \\  \hline \hline
	\end{tabular}}
	\caption{Comparison of the proposed method with existing NMT systems on the En-De, Ja-En, and Zh-En translation tasks. ``\#Speed." and ``\#Para." denote the training speed (tokens/second) and the size of model parameters, respectively. ``\(\Uparrow\)" after the score indicates that the proposed method was significantly better than the corresponding baseline Trans.base/big at significance level p$<$0.01. \textbf{Note that we did not use the target test data set and \textit{fast\_align} as in the preliminary experiment in Section~\ref{sec3_PreliminaryExperiment}. }}
	\label{Tab1:MainResults}
\end{table*}
Table~\ref{Tab1:MainResults} lists the translation results of the three translation tasks (including En-De, Ja-En, and Zh-En) and other existing comparison systems.

\noindent \textbf{Main Results}:
Our re-implemented Trans.base/big models outperform the reported results in the original Trans.base/big ~\cite{NIPS2017_7181} for the same En-De dataset, which makes the evaluation convincing.
As seen, in terms of BLEU score, the proposed +ExGRE and +ReFSR consistently improve translation performance of Trans.base/big models on the En-DE, Ja-En, and Zh-En tasks, which demonstrates the effectiveness and universality of the proposed approach.

\noindent \textbf{Effect of Target Order}: Trans.base/big+ExGRE (28.44/29.28) outperformed the existing works related with positional information, for example, +Relative PE (26.8/29.2), +Reordering PE (28.22/29.11), +Recurrent PE (28.35/29.11), +structural PE (*/28.88), +CSANs (28.18/28.74), and +BIARN (28.21/28.98) for the En-DE task. 
This means that target order information is a useful translation knowledge, and the proposed ExGRE can better model it to improve the performance of the Transformer-based NMT model.  

\noindent \textbf{Fusion Order Evaluation}:
Trans.base/big +ExFSR gains the highest BLEU score among all of the methods, including +Relative PE, +Reordering PE, +Recurrent PE, +structural PE, +BIARN, and +CSANs.
In particular, Trans.base/big+ExFSR (28.65/29.54) obtain further improvements over the Trans.base/big+ExGRE (28.44/29.28).
This indicates that our learned target order information can be combined with the original order information to further improve translation performance.

\noindent \textbf{Model Parameters and Training Speed}:
Compared with the baseline and comparison methods, both +ExGRE and +ReFSR  nearly don't introduce additional model parameters.
This means that the performance improvement is indeed from the proposed method rather than model parameters.
The training speed of the proposed +ExFSR decreased (10\%) because of encoding two types of sequences, compared to the corresponding baselines.
In addition, the \textit{fast\_align} toolkit took 0.4, 0.2, and 1.1 hours to gain word alignments  for the En-De, Ja-En, and Zh-En training datasets.

\noindent \textbf{Universality of Our Method}:
In addition to the En-De task, the proposed +ExGRE and +ExFSR models yielded similar improvements over the baseline system and the compared methods on the Ja-En and Zh-En translation tasks.
This means that our method is a universal method for improving the translation of other language pairs.

\subsection{Evaluating Explicit Reordering Embeddings}
\begin{figure}[h]
		\begin{center}
				\pgfplotsset{height=5.0cm,width=7.0cm,compat=1.14,every axis/.append style={thick},legend columns=1}
				\begin{tikzpicture}
				\centering
				\begin{axis}[
				ybar=5pt,
				bar width=8pt,
				enlargelimits=0.15,
				legend style={at={(0.78,0.95)},
					anchor=north, legend columns=-1},
				ylabel={Sim},
				font=\small,
				symbolic x coords={En-De, Ja-En, Zh-En},
				xtick=data,
				bar width=9pt,
				],
				
				\addplot [draw=blue]
				coordinates 
				{(En-De,0.767) (Ja-En,0.532) (Zh-En,0.614)};
		        \addplot [draw=red]
		        coordinates 
		        {(En-De,0.844) (Ja-En,0.613) (Zh-En,0.683)};
				\legend{PE,PR}
				\end{axis}
				\end{tikzpicture}
		\end{center}
	\caption{Similarity score between our reordered positional embeddings and the supervised positional embeddings for the Trans.base+ReRSR model.}
	\label{fig5:PRSimilar}
\end{figure}
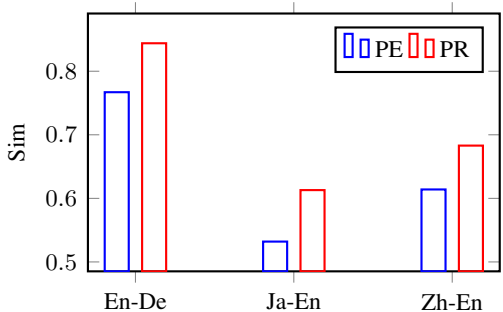

To evaluate the explicit reordering \(\textbf{PR}^{N}\) in Eq.\eqref{eq7:Reordered_Embeding}, we computed the averaged cosine distance between our reordered positional embedding \(\textbf{pr}^{N}_j\) and the supervised positional embedding \(\textbf{re}_j\) on the test set:
\begin{equation}
	\textup{Sim} = \frac{1}{\sum_{k=1}^{K}L_{k}}\sum_{k=1}^{K}\sum_{j=1}^{L_k}\frac{\textbf{pr}^{N}_j\cdot\textbf{re}_j}{\parallel \textbf{pr}^{N}_j \parallel\parallel \textbf{re}_j\parallel},
	\label{eq12:Semantic_Similarity}
\end{equation}
where $K$ and $L_K$ are the total sentence numbers of test set and the length of the \(K\)-th sentence, respectively. 
Figure~\ref{fig5:PRSimilar} presents the computed averaged similarity score on the En-De, Ja-En, Zh-En test sets. 
As seen, the proposed reordered PE is closer to the supervised PE learned from word alignments.

\subsection{Evaluating Reordering Loss}
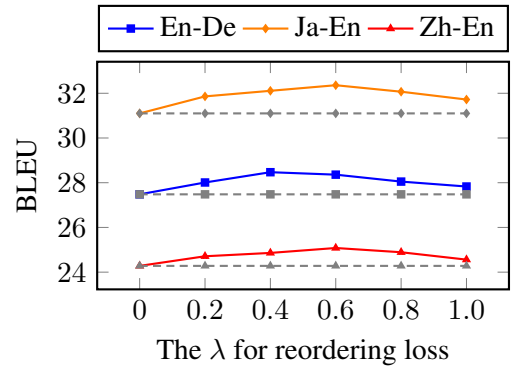
\begin{figure}[h]
	\begin{center}
		\pgfplotsset{height=4.6cm,width=7.0cm,compat=1.14,every axis/.append style={thick},legend columns=3}
		\begin{tikzpicture}
		\tikzset{every node}
		\begin{axis}
		[width=7.0cm, enlargelimits=0.13, tick align=inside,every axis legend/.append style={
			at={(1.0,1.25)}}, xticklabels={$0$, $0.2$,$0.4$, $0.6$, $0.8$, $1.0$},
		xtick={0,1,2,3,4,5},
		x tick label style={rotate=0},
		ylabel={BLEU},xlabel={The \(\lambda\) for reordering loss}]
		
		\addplot+[sharp plot, mark=square*,mark size=1.2pt,mark options={solid,mark color=blue}, color=blue] 
		coordinates
		{(0,27.48) (1,28.01) (2,28.47) (3,28.36) (4,28.05) (5,27.83)};
		\addlegendentry{En-De}
		
		\addplot+[sharp plot, mark=diamond*,mark size=1.2pt,mark options={solid,mark color=orange}, color=orange] 
		coordinates
		{(0,31.10) (1,31.86) (2,32.11) (3,32.36) (4,32.07) (5,31.72) };
		\addlegendentry{Ja-En}
		
		\addplot+[sharp plot, mark=triangle*,mark size=1.2pt,mark options={solid,mark color=red}, color=red] 
		coordinates
		{(0,24.28) (1,24.71) (2,24.86) (3,25.08) (4,24.89) (5,24.56) };
		\addlegendentry{Zh-En}
		
		\addplot+[sharp plot, densely dashed, mark=square*,mark size=1.2pt,mark options={solid,mark color=gray}, color=gray] 
		coordinates
		{(0,27.48) (1,27.48) (2,27.48) (3,27.48) (4,27.48) (5,27.48)};
		
		\addplot+[sharp plot, densely dashed, mark=diamond*,mark size=1.2pt,mark options={solid,mark color=gray}, color=gray] 
		coordinates
		{(0,31.1) (1,31.1) (2,31.1) (3,31.1) (4,31.1) (5,31.1) };
		
		\addplot+[sharp plot, densely dashed, mark=triangle*,mark size=1.2pt,mark options={solid,mark color=gray}, color=gray] 
		coordinates
		{(0,24.28) (1,24.28) (2,24.28) (3,24.28) (4,24.28) (5,24.28) };
		
		\end{axis}
		\end{tikzpicture}
		\caption{BLEU scores of the Trans.base+ReFSR model on the three test sets with different values of \(\lambda\). The gray dashed line denotes the result of the Trans.base model}
		\label{fig6:Reordering_loss_hyper_parameter}
	\end{center}
\end{figure}

Figure~\ref{fig6:Reordering_loss_hyper_parameter} shows the results of Trans.base+ReFSR model on the En-De, Ja-En, and Zh-En test sets with different hyper-parameter $\lambda$ for reordering loss.
When $\lambda$ is one of (0.2, 0.4, 0.6, 0.8), the result of Trans.base+ReFSR model outperformed the Trans.base on the three test sets.
Furthermore, Trans.base+ReFSR model reached the point of highest BLEU score with $\lambda$ increasing from 0 to 0.6 whereas the point of highest BLEU score is $\lambda$=0.4 on the En-De test set.
We think that there are larger word order difference between distant language pair (i.e., Ja-En and Zh-En) than similar language pair (i.e., En-De).
\begin{figure}[h]
	\begin{center}
		\pgfplotsset{height=5.3cm,width=7.0cm,compat=1.14,every axis/.append style={thick},legend columns=1}
		\begin{tikzpicture}
		\begin{axis}
		[ybar stacked, enlargelimits=0.13, tick align=inside,every axis legend/.append style={
			at={(1.0,1.4)}}, anchor=north, xticklabels={Ar-Fr, En-Fr, Es-Fr, Ru-Fr, Zh-Fr},
		xtick={0,1,2,3,4},
		x tick label style={rotate=0},
		ylabel={BLEU scores},xlabel={Source languages}]
		
		\addplot
		coordinates
		{(0,48.00) (1,53.68) (2,51.86) (3,47.26) (4,44.29)};
        
  	    \addplot
  	    coordinates
		{(0,0.96) (1,0.48) (2,0.65) (3,1.05) (4,1.14)};
		
		\end{axis}
		\end{tikzpicture}
	\end{center}
	\caption{Results of the Trans.base+ReFSR on Ar-Fr, En-Fr, Es-Fr, Ru-Fr, and Zh-Fr datasets from the UNv1.0 six language parallel corpus. Red area denotes the increasing improvement of +ReFSR over the Trans.base model (blue area).}
	\label{fig7:Effect_Different_Language_Pair}
\end{figure}
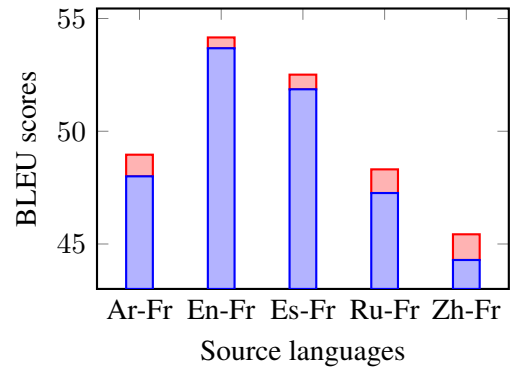
\subsection{Effect of Different Language Pairs}

To evaluate effect of different language pairs, we applied the proposed method to six language parallel corpus from the UNv1.0 Parallel Corpus.
In detail, the french language was selected as target language and other five languages, were served as the source language, and there were five including Arabic-to-French (Ar-Fr), English-to-French (En-Fr), Spanish-to-French (Es-Fr), Russian-to-French (Ru-Fr), and Chinese-to-French (Zh-Fr) translation datasets.
Figure~\ref{fig7:Effect_Different_Language_Pair} showed BLEU scores of the Trans.base model and +ReFSR model on five language pairs.
AS seen, the improvements of Ar-Fr, Ru-Fr, and Zh-Fr language pairs were greater than that of En-Fr and Es-Fr language pairs.
Since the proposed method focused on adapting the word order of the source sentence to that of the target sentence, We think that one of the reasons may be that there are a larger word order differences in Ar-Fr, Ru-fr, and Zh-Fr language pairs than in En-Fr and Es-Fr language pairs.
In other words,  the proposed method is more effective in distant language pairs (i.e., Ar-Fr, Ru-fr, and Zh-Fr) than similar language pairs (i.e., Es-Fr and En-Fr).

\subsection{Translation Case}
\begin{figure*}[t]
	\centering
	\includegraphics[width=6.1in, height=1.4in]{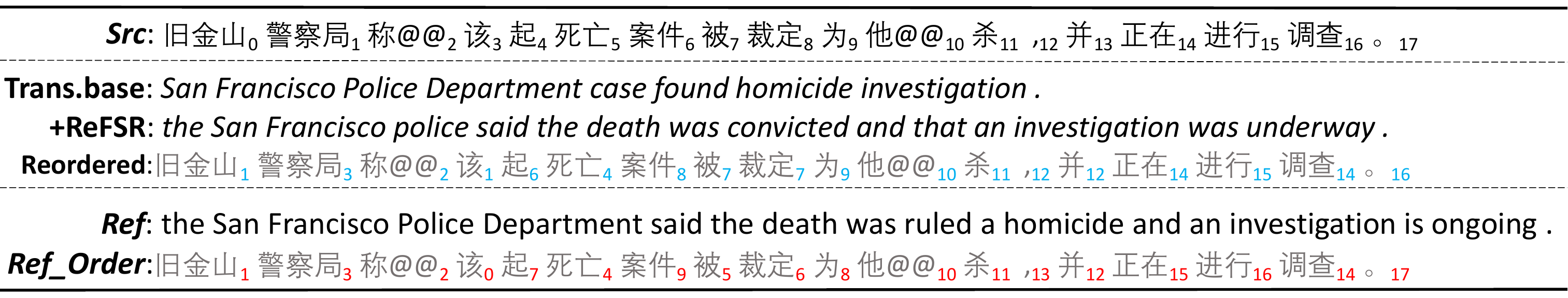}
	\caption{A Zh-En translation case. The blue and red subscripts of ``num" denote \(s\)\(\in\)\([b_{j}^{n}-D, b_{j}^{n}+D]\) in Eq.\eqref{eq7:Reordered_Embeding} and the supervised word order learned from word alignments, respectively.} 
	\label{fig8:Translation_Case}
\end{figure*}
Figure~\ref{fig8:Translation_Case} shows a case of reordering and translation on the Zh-En translation task.
We observe that the result of +ReFSR is closer to that of the reference translation ``\textbf{\textit{Ref}}".
In particular, most of the ``\textbf{Reordered}" are consistent with the supervised ``\textbf{\textit{Ref}}\_\textbf{\textit{Order}}".
Besides, there are also some repeated positions, for example, position one, seven, twelve, and fourteen in ``\textbf{Reordered}".
We think that this is one of reasons why the performance of the proposed method failed to achieve so impressive performance as using supervised alignment in the test data which is shown in Figure~\ref{fig2:PreliminaryExperiments} (a). 

\section{Related Work}
In traditional phrase-based SMT, There are many works for modeling reordering information, including phrase orientation models~\cite{P06-1090,D08-1089}, jump models~\cite{P06-1067,N10-1129}, source decoding sequence models~\cite{feng2010source,P13-1032}, operation sequence models~\cite{P11-1105,P13-2071}, and ITG-based reordering models~\cite{D13-1054,C14-1179}.
Typically, these works learn large-scale reordering rules from parallel bilingual sentence pairs in advance to ensure fluent translations.
However, these statistical-based reordering rules are generally difficult to be compatible with sequence-to-sequence NMT~\cite{sutskever2014sequence,bahdanau2015neural,NIPS2017_7181}.

Inspired by the distortion mechanism~\cite{N03-1017,P06-1067} of SMT, \newcite{P17-1140} introduced position-based attention to model word reordering penalty for the traditional RNN-based NMT model.
For state-of-the-art Transformer-based NMT model, \newcite{shaw-etal-2018-self} proposed to a relative position representation to encode order information in a sentence instead of the existing absolute position representation.
\newcite{ma-etal-2019-improving} and \newcite{wang-etal-2019-self} introduced syntax distance constraints of the dependency tree to model order dependency between words for the Transformer-based NMT.  
Recently, \newcite{chen-etal-2019-neural} proposed a reordering embedding to implicitly model reordering information to improve the translation performance of Transformer-based NMT.

Moreover, \newcite{P18-3004} pre-reordered the word orders of source to those of the target as the input to the existing RNN-based NMT, but reported a negative result for the WAT ASPEC Ja-En translation task.
\newcite{P18-3004} assumed that one reason is the isolation between pre-ordering and NMT models. 
This paper first empirically found that source input with the order learned from the parallel target language can substantially improve translation performance.
Thus, we propose a novel reordering method to explicitly model this reordering information to improve Transformer-based NMT.

\section{Conclusion and Future Work}
In this paper, we empirically find a useful target order information in the source sentence for Transformer-based NMT.
Based on this impressive finding, we extract this useful target order information from bilingual sentence pair through word alignment to explicitly model the source reordering knowledge. 
Thus, we proposed two simple and efficient methods to integrate this reorder information into Transformer-based NMT model for enhancing translation predictions.
Experiments on three large-scale translation tasks shows that the proposed method can significantly improve translation performance. 
In future work, we will explore an more effective method to fully model the target order information. 

\bibliographystyle{acl_natbib}
\bibliography{emnlp2020}
\end{document}